%% file: acl_latex.tex
\pdfoutput=1

\documentclass[11pt]{article}

\usepackage[final]{acl}

\usepackage{times}
\usepackage{latexsym}
\usepackage{xspace}

\usepackage[T1]{fontenc}

\usepackage[utf8]{inputenc}

\usepackage{microtype}

\usepackage{inconsolata}

\usepackage{graphicx}
\usepackage{amsmath}
\usepackage{amssymb}
\usepackage{algorithm}
\usepackage{algpseudocode}
\usepackage{booktabs, tabularx}
\usepackage{xcolor}
\title{Consistent Document-Level Relation Extraction via Counterfactuals}

\author{
    Ali Modarressi$^{1,2}$ ~ Abdullatif Köksal$^{1,2}$ ~ Hinrich Schütze$^{1,2}$\\
    $^1$Center for Information and Language Processing, LMU Munich, Germany \\
    $^2$Munich Center for Machine Learning, Germany \\
    \texttt{amodaresi@cis.lmu.de}
}

\newcounter{notecounter}
\newcommand{\enotesoff}{\long\gdef\enote##1##2{}}
\newcommand{\enoteson}{\long\gdef\enote##1##2{{
			\stepcounter{notecounter}
			{\large\textbf{ \hspace{1cm}\arabic{notecounter} $<<<$ ##1: ##2 $>>>$\hspace{1cm}}}}}}
\enoteson
\enotesoff

\newcommand{\alidata}{\textsc{Re-DocRED-CF}\xspace}
\newcommand{\alipipeline}{\textsc{CovEReD}\xspace}

\begin{document}
\maketitle
\begin{abstract}
Many datasets have been developed to train and evaluate
document-level relation extraction (RE) models.  Most of
these are constructed using real-world data. It has been
shown that RE
models trained on real-world data
suffer from factual biases. To evaluate
and address this issue, we present
\alipipeline,
\textbf{a counterfactual
data generation approach for document-level relation
extraction datasets using entity replacement}. We first
demonstrate that models trained on factual data exhibit
inconsistent behavior: while they accurately extract
triples from factual data, they fail to extract the same
triples after counterfactual modification.  This
inconsistency suggests that models trained on factual data
rely on spurious signals such as specific entities and
external knowledge -- rather than on the input context -- to extract
triples.  We show that by generating document-level
counterfactual data with \alipipeline and training models on them, consistency
is maintained with minimal impact on RE performance.  We
release our \alipipeline pipeline\footnote{\url{https://github.com/amodaresi/CovEReD}}
as well as \alidata, a dataset of counterfactual RE documents,
to assist
in evaluating and addressing inconsistency in document-level
RE.
\end{abstract}


\enote{hs}{the abstract doesn't explain what consistency is
and it doesn't sell the apper very well: why is consistency important?}

\enote{hs}{the figure on page 1 doesn't sell the paper very
well. who is interested in an IE system getting a
counterfactual extraction right? it would be better to have
a figure that shows the benefit for information extraction
that the new method brings. basically our use case in the
memory project: only a counterfactually trained model
develops good IE capabilitesi beyond what's in its
parametric memory}

\enote{ak}{i think it kinda states that factual models have
this `factuality' bias, and can't understand the context, decide
only based on entities. so, i like the examples but it can be
explained better by saying that the factual model probably 
just memorized entities. (ofc it'd be better if there's an example
like hinrich suggested, then this example can go to appendix or
somewhere else in the paper.)}

\enote{hs}{i didn't undderstand Algorithm 2. At a minium,
can you please explain what EditTuple does / what role it has?}

\enote{ak}{you can explain factual bias like stating that
these models learn spurious correlations based on entities.
then it would be easier to explain why consistency is important,
because it shows that the model didn't learn just based on memorizing 
entities.}

\section{Introduction}

\input{figures/CF_example_1.tex}
Relation extraction (RE)  extracts triples,
semantic relations between two entities, from text.
In document-level RE, triples can span multiple
sentences \citep{yao-etal-2019-docred,
tan-etal-2022-revisiting,
xiaoyan2023comprehensive}. RE datasets such as
DocRED \citep{yao-etal-2019-docred} and
Re-DocRED \citep{tan-etal-2022-revisiting}
consist of a factual corpus (Wikipedia) annotated with triples.
Most recent DocRE models are
based on pretrained language models (PLMs) \citep{tang2020hin,
zhou2021document, tan-etal-2022-document}
trained on these datasets. While PLMs
perform strongly,
they are susceptible to
factual biases and other spurious correlations.
To generate
triples, 
instead of inferring from the input,
they may use their parametric knowledge \citep{mccoy-etal-2019-right, Kaushik2020Learning,
paranjape-etal-2022-retrieval}.  A common case is entity bias:
the model relies on entities
in its parametric knowledge to make a
prediction \citep{longpre-etal-2021-entity,
qian-etal-2021-annotation,
xu-etal-2022-model, chen-etal-2023-models}.

\citet{wang-etal-2022-rely} perform a
counterfactual analysis (CoRE) for sentence-level RE. They
remove the context and provide only the entity
mentions. They then distil the biases and
propose a debiasing method using a causal
graph. ENTRE \citep{wang-etal-2023-fragile}, a
counterfactual modification of
TACRED \citep{zhang-etal-2017-position},
replaces entities to develop a robust
sentence-level
RE benchmark. They show that RE models rely on
memorized facts instead of the sentence context.
All of this work is focused on
\emph{sentence-level} RE.

This paper presents \alipipeline, a counterfactual (CF) data generation
method for \emph{document-level} RE. It
replaces entities and thereby generates text
containing triples with
minimal factual alignment.
We apply \alipipeline to Re-DocRED, creating \alidata, 
a counterfactual document-level RE dataset.
Since we apply replacements on the document level,
our method handles multiple entity mentions and also
multiple replacements at a time --
unlike sentence-level
methods.
We achieve this by
considering all triples that an entity is
involved in and embeddings of its contexts.
Evaluation on \alidata
allows us to measure how consistent a model is in RE.
We show that models trained on factual
documents lack robustness against nonfactual data
(Figure \ref{fig:cf_example_main}).
We then train an RE model on Re-DocRED and \alidata and
show that it has
high consistency with only a negligible effect on 
accuracy on factual data.
Our approach is novel in that it creates counterfactual
datasets on the document level -- the level at which RE is
used in a real application --
to analyze and improve DocRE
models. Alongside \alipipeline, the data generation pipeline, we 
release \alidata, a counterfactual dataset
generated from Re-DocRED.


\enote{ak}{It may be better to mention that `we are releasing a
	counterfactual RE dataset publicly with xxxx number of examples'.}

\enote{hs}{i don't think you state the goal of what you're
doing in this section. you want to improve the IE capability
of an LLM through counterfactual trainign. however, if you
use wildly implausible counterfactual facts for training,
then this won't work. so what you do in this section is to
make sure that the counterfactual facts you produce are
relatively plausible, so that yuo then have good training amterial}

\section{Counterfactual pipeline and dataset}
To evaluate and address robustness against factuality bias,
we need to generate documents
from which such biases have been removed.
Hence, in this section we describe \alipipeline, our
mechanism for generating counterfactual (CF) documents from a 
document-level RE dataset. Our seed dataset is
Re-DocRED  \citep{tan-etal-2022-revisiting}.
For each
Re-DocRED document
$d$, we
have a set of entities $E$ and a set of relation triples
$R$. For each entity node $e_i \in E$, the dataset provides
the positioning of each mention of $e_i$ and its
type (ORG, TIME \ldots). In a triple $r \in R$, we have the
indices ($i$) of head and tail entities, the relation
$r_t$ and -- if the triple comes from the original DocRED
\cite{yao-etal-2019-docred} -- the
IDs of the sentences that are the evidence for
$r$.

\enote{hs}{above: the indices of head and tail can be
different, correct?}

To generate counterfactuals from Re-DocRED's documents,
our pipeline \alipipeline
proceeds with the following three steps (\S\ref{cleanup}--\S{\ref{generate}).

\subsection{Entity mention cleanup}
\label{cleanup}
If two entities $e_i$ and $e_j$ share a common (exactly matching) mention,
then we merge them; this means that we merge the two sets of
mentions and treat $e_i$ and $e_j$ as synonymous.
Also, if two mentions overlap in a sentence, we
discard the shorter one and
only keep
the longer one. 
Example: If ``Great Britain'' in a sentence is
annotated with two mentions ``Great Britain'' and
``Britain'', then we only keep ``Great Britain''.


\subsection{Gathering entity candidates}
\label{sec:gatheringentitycandidates}
In sentence-level RE,  entities are very rarely part of multiple
triples, but 
in document-level RE this is common.  If we want to generate consistent
counterfactual documents, we have to replace all of an entity's
occurrences.  This makes it impractical to use
simplistic replacement methods in document-level RE -- such as relying
on an entity bank for random replacement as  in
ENTRE \cite{wang-etal-2023-fragile}.

Another challenge is that we need ``plausible''
counterfactual documents
that do
not obviously contradict general knowledge. For example,
``Obama was born in Panthera leo'' (where Panthera leo is a
species, the lion) is too implausible to teach the model about correct
RE. We therefore only replace $e_i$ with $e_j$ if  they
have (i) similar \textbf{relation
maps} and
(ii) similar \textbf{context snippets}.
For this step, we use the set of
entities over the entire seed dataset.



The \textbf{relation map} for an entity $e_i$ is a set of pairs, each
consisting of a relation and the position of $e_i$ within
that relation (head or tail).
For example, the relation map of
``United
States'' -- occurring in  triples such as
$\langle \text{NBA}, \text{country}, \text{United
States}\rangle$ -- may contain 
the pair $(\text{country}, \text{tail})$.
If two entities $e_i$ and $e_j$ have 
similar relation maps, then
$e_i$ is a 
good candidate
for replacing $e_j$ since they occur in similar triples.

The \textbf{context snippet} of a mention $m$
includes up to
16 words on each side of $m$.
For each context snippet, we compute its embedding
(using  Contriever
 \cite{izacard2022unsupervised}).
If two entities $e_i$ and $e_j$ have 
similar context embeddings (as measured by cosine similarity), then
$e_i$ is a 
good candidate
for replacing $e_j$ since they occur in similar contexts.

\subsection{Generating counterfactual documents}
\label{generate}
Our general approach to generating counterfactual documents is to find
suitable entity alternatives for each entity node and apply
replacements.


In Algorithm \ref{algo:cfGenerator}, function
\textsc{GetAlts} is responsible for finding suitable entity
replacements. For each entity node, we compare its
features (its type, relation map, mention and context
snippet embeddings; see \S\ref{sec:gatheringentitycandidates}) with
the \emph{candidates} -- the other entity nodes in the
pool $E$ gathered from the document collection.
We deem a candidate  a suitable alternative
if it  is similar and not from the
same document. 
We do not want candidates's entity mentions
(as measured by cosine similarity of the embeddings of the
entity mention strings)
to be too  similar;
e.g., ``United States'' vs
``U.S.''. The reason is that we of course want
the candidate to be a different entity.
\textsc{GetAlts}  returns a list of sets, each containing,
for a particular alternative entity $e_i$, 
possible mention strings for $e_i$. For
instance, if the entity node we want to replace is
``United States'', an example mention set is
$\{$``United Kingdom'', ``UK'', ``Britain''$\}$. For
more details
see
Appendix \ref{apdx:getAlternatives}.

To generate counterfactual documents from the seed document $d$, we
loop over all entity nodes in $d$.
We attempt to apply replacements for each entity node.
To achieve this, we
first create an empty dictionary
-- denoted as $\mathbb{D}$ --
for our newly generated
documents (each created through replacements).
After having replaced an entity node,
we add the resulting counterfactual document to this
dictionary.
We use
EditTuple to record which nodes have been replaced,
preventing any node from being replaced more than once.
We repeatedly loop over the
dictionary to gather a large number of counterfactual documents.
After the replacement process is completed,
we select those counterfactual documents that are
affected by the replacement more than a  threshold
$\tau_r$. Thus, we require that a valid counterfactual document have at
least $\tau_r$ percent of their triples altered (in either
one or both entities).


\enote{hs}{is this case taken care of in the pseudocode?
  myinterpretration is : no, it isn't
but \textsc{GetAlts} can return nothing as
an alternative, that is why we try out every combination.
}

\begin{algorithm}[!ht]
  \small
  \caption{Counterfactual Example Generator}
  \begin{algorithmic}[1]
    \Statex \textbf{Input:}
    \Statex \hspace{1em} $d$: Document with entity nodes and relation triples
    \Statex \hspace{1em} $\tau_r$: Affected relations threshold---An augmented document should have more than $\tau_r$ of its relations affected by the replacements
    \Statex \hspace{1em} $M_N$: Maximum number of alternatives to sample from
    \Statex \textbf{Output:}
    \Statex \hspace{1em} $\mathcal{D}$: A set of documents with entity replacements applied on $d$
    \Statex \textbf{Auxiliary functions:}
    \Statex \hspace{1em} \textsc{Replace}($i$, $d$, $alt$): A function that replaces node ($i$) and its mentions in the document ($d$) with a given alternative entity mentions set ($alt$).\footnotemark
    \Statex \hspace{1em} \textsc{AffectR}(EditTuple, $d$): This function returns the ratio of relation triples that are affected by the replacements specified in the EditTuple.
    \Statex \hspace{1em} \textsc{GetAlts}($e_i$,$\mathbb{E}$,$\tau_\text{e[MAX]}$,$\tau_\text{e[MIN]}$,$\tau_\text{c}$): This function returns a list of sets of alternatives for the given entity node $e_i$. It requires the candidates pool $\mathbb{E}$, and other sets of hyperparameters (cf. Appendix \ref{apdx:getAlternatives}).
    \Statex
    \State Initialize $\mathbb{D} \gets \{\}$, EditTuple $\gets ()$, $\mathcal{D} \gets [\ ]$
    \State $\mathbb{D}[\text{EditTuple}] \gets d$
    \For{\text{EditTuple}, $\tilde{d}$ in $\mathbb{D}$}
      \For{$e_i$ in $\tilde{d}[\text{EntityNodes}]$}
        \If{$i \notin \text{EditTuple}$}
          \State $alts \leftarrow$ \Call{GetAlts}{$e_i$,$\mathbb{E}$,$\tau_\text{e[MAX]}$,$\tau_\text{e[MIN]}$,$\tau_\text{c}$}
          \State Sample $alt$ from $alts[:M_N]$
          \State Add $i$ to \text{EditTuple}
          \State $\mathbb{D}[\text{EditTuple}] \gets$ \Call{Replace}{$i$,$\tilde{d}$,$alt$}
        \EndIf
      \EndFor
    \EndFor
    \For{\text{EditTuple}, $\tilde{d}$ in $\mathbb{D}$}
      \If{\Call{AffectR}{EditTuple, $d$} > $\tau_r$}
        \State Add $\tilde{d}$ to $\mathcal{D}$
      \EndIf
    \EndFor
    \State \Return $\mathcal{D}$
  \end{algorithmic}
  \label{algo:cfGenerator}
\end{algorithm}
\footnotetext{For each replacement the closest mention to the original mentions, in terms of embedding similarity would be selected.}

\section{Experiments}
We generate
\alidata, 
our counterfactual dataset, from Re-DocRED
using \alipipeline.
We run \alipipeline five times on Re-DocRED train
to produce \alidata train (so it consists of
five different counterfactual datasets).
We run \alipipeline  on Re-DocRED test once to generate
\alidata test.
We
set $\tau_\text{e[MAX]}=.8$, $\tau_\text{e[MIN]}=.2$,
$\tau_\text{c}=.4$, $M_N= 3$ (to limit the search)
and  (to
make sure at least 70\% of triples are affected by the
replacements) $\tau_r=.7$.

We first evaluate the hypothesis that models that are trained
merely on factual data do not reliably use the context for
the RE task. To test this, we measure how 
consistent these models are for documents that have undergone entity
replacement.
We use the KD-DocRE framework
\citep{tan-etal-2022-document}\footnote{https://github.com/tonytan48/KD-DocRE}
to train DocRE models. The framework features axial
attention modules, adaptive focal loss and knowledge
distillation over the distant supervised examples. As we
want to observe the effect of using counterfactual data in
training, we do not use  knowledge distillation and only do their first stage of training over the human-annotated data.

We follow \citet{tan-etal-2022-document}'s setup and
hyperparameters in finetuning a RoBERTa-large
model \citep{liu2019roberta} for relation extraction. First,
with the training set of Re-DocRED, we finetune a model that
we probe for factual biases. To mitigate random errors, we
train with five random seeds and report the median over each
metric.

\enote{ak}{as hinrich suggested above, you can also give a
  motivation here.  why is pairwise consistency important?}

To assess a model's factual bias, we need to observe how its
behavior changes when presented with counterfactual data.
Following \citet{paranjape-etal-2022-retrieval}, we use
\emph{pairwise consistency} as our measure.  Pairwise
consistency is the accuracy
of the model on those counterfactuals whose factual
counterparts (the original facts) were predicted correctly.

\subsection{Results}
Table \ref{tab:exp_results} shows performance of the trained
models on Re-DocRED test.  As expected, the
model that is only trained on factual data performs well on
similar factual data. However, it only shows 68.6\%
consistency
(for Re-DocRED test and \alidata test) --
more than 30\% of the correctly
predicted output is based on entity and factual biases. 
Figure \ref{fig:cf_example_main} shows
the original text (top) and the text after
replacement of entities (bottom). 
We see that \alipipeline, our replacement algorithm, did a good job
here: the original entities were replaced with similar
entities (which occur in similar contexts and with similar
relations), but of course the new triples are nonfactual. 
The model
correctly predicted the triple (Cleanin' Out My
Closet, part of, The Eminem Show) from the original
document. However,
for the document with replaced entities
-- even though 
the relation (``part of'')
is still the same, only the entities have changed --
the model fails to extract the correct
triple,
which would be:
(The Ultimate Collection, part of,
London Calling).
See Appendix \ref{apdx:extra_examples} for more examples.  This
result corroborates other similar analysis that was
done on the sentence level \citep{wang-etal-2022-rely,
  wang-etal-2023-fragile}.  \input{tables/results.tex}

To evaluate the effectiveness of \alipipeline in generating plausible examples, we conducted a human evaluation on a sample subset of the data. We randomly selected 50 triplets from the test set and found that 45 of them were deemed plausible. This indicates that 90 percent of the counterfactual triplets accurately reflect relationships that are evident from the counterfactual version of the document.\footnote{For this evaluation, we excluded counterfactual examples where the original counterparts were already mislabelled (e.g., due to entity linking errors or misannotation of non-evident relations). This ensures that our analysis focuses on evaluating the plausibility of our pipeline on correctly labeled examples.}

Our hypothesis is that we can increase robustness against
entity and factual biases by training on counterfactual
data.
We
finetune a separate model with its own separate random seed
for each of the five parts of \alidata train.
Table \ref{tab:exp_results} shows that
consistency  increases with a
$>$20\% gap compared to only using factual data
(89.5\% vs 68.6\%).  This shows that counterfactuals improve
the model's robustness against entity and factual biases. However,
training on counterfactuals only also deteriorates
performance on factual test data (5.6 drop on F1, 72.4 vs
78.0). Since the real-world use case of DocRE models is
factual data, we need to devise a solution that is  both
performant and consistent.

Therefore, we conduct a third experiment in which we mix
each of the five subsets of \alidata train with Re-DocRED train.
To keep the number of training steps equal, we halve the number
of epochs of training that we used in the other two experiments
($30 \rightarrow 15$). As shown in Table
\ref{tab:exp_results}, the resulting model shows both a high
performance with minimum drop in F1 (only  -1.7, 76.3 vs 78.0) while
also being consistent (88.3\% vs 68.6\% for the
``factual-training-only'' model).  This means that the
counterfactual \alidata dataset helps the model to learn the task based
on the context and mitigates bias issues while having the
factual dataset alongside keeps the model performant on
factual data.

\section{Conclusion}
In this work, we present a method for generating
counterfactual examples for document-level relation
extraction. Our approach searches for suitable entity
replacements over a document and applies them to a point where
most of the relations are affected by these replacements. By
generating a counterfactual test set, we demonstrate the
high level of inconsistency DocRE model  have when
trained only with factual data. Adding counterfactuals to
the training sets improves consistency  by a
large margin while keeping performance  high.
We make our pipeline \alipipeline and dataset \alidata
publicly available
We
hope our findings and resources will raise awareness
and support future efforts in addressing entity and factual biases in
document relation extraction.

\enote{ak}{in addition to increasing awareness, you are also releasing a dataset,
mentioning that could be a good idea here to facilitate future work.}

\section*{Limitations}
The main limitation of this work is its requirement of a seed DocRE dataset. This means to extend this approach to either other domains or languages we need a document RE dataset provided. Here, we measured consistency and performance levels on KD-DocRE, one of the recent and high-performing methods. However, other solutions might yield different performance results. Our aim in this work is to provide a document-level RE dataset for consistency. Also, the improvements in robustness against factual bias were gained in an equal setup.

\section*{Acknowledgements}
This work was funded by Deutsche Forschungsgemeinschaft
(Project SCHU 2246/14-1).

\bibliography{custom}

\appendix
\input{figures/CF_example_2.tex}
\section{Alternative Entities Search Algorithm}
\label{apdx:getAlternatives}
Algorithm \ref{algo:getAlternatives} is the detailed pseudocode of our approach in finding suitable alternatives for an entity node.
\begin{algorithm}[!ht]
  \small
  \caption{Find Suitable Alternatives for a given Entity Node $e_i$}
  \begin{algorithmic}[1]
    \Statex \textbf{Input:}
    \Statex \hspace{1em} $e_i$: Input entity node
    \Statex \hspace{1em} $\mathbb{E}$: Entity candidates pool
    \Statex \hspace{1em} $\tau_\text{e[MAX]}$, $\tau_\text{e[MIN]}$: Maximum and minimum entity mention similarity threshold
    \Statex \hspace{1em} $\tau_\text{c}$: Context similarity threshold
      \Statex \textbf{Output:} 
      \Statex \hspace{1em} $\mathcal{E}$: List of sets of alternative entity mentions for the given entity node
      \Function{GetAlts}{$e_i$,$\mathbb{E}$,$\tau_\text{e[MAX]}$,$\tau_\text{e[MIN]}$,$\tau_\text{c}$}
      \State Initialize: $\mathcal{E} \gets [\ ]$, $\mathcal{E}' \gets [\ ]$
      \For{$\tilde{e}$ in $\mathbb{E}$}
          \State $r_\text{sim} = |\tilde{e}[\text{rel\_maps}] \cap e_i[\text{rel\_maps}]|$
          \If{$r_\text{sim}= 0$}
            \State \textbf{continue}
          \ElsIf{$\tilde{e}[\text{doc\_title}] = e_i[\text{doc\_title}]$}
            \State \textbf{continue}
          \ElsIf{$\tilde{e}[\text{type}] \cap e_i[\text{type}] = \varnothing$}
            \State \textbf{continue}
          \EndIf
          \State Set: $m_\text{sim} \leftarrow 0$
          \For{$m_i$ in $e_i[\text{mentions}]$, $\tilde{m}$ in $\tilde{e}[\text{mentions}]$}
            \State $\text{sim} = \cos(\tilde{m}[\text{emb}],m_i[\text{emb}])$
            \State $m_\text{sim} \leftarrow \max(m_\text{sim}, \text{sim})$
          \EndFor 
          \State Set: $c_\text{sim} \leftarrow 0$
          \For{$c_i$ in $e_i[\text{contexts}]$, $\tilde{c}$ in $\tilde{e}[\text{contexts}]$}
            \State $\text{sim} = \cos(\tilde{c}[\text{emb}],c_i[\text{emb}])$
            \State $c_\text{sim} \leftarrow \max(c_\text{sim}, \text{sim})$
          \EndFor
          \If{$\tau_\text{e[MIN]} < m_\text{sim} < \tau_\text{e[MAX]}$ \textbf{and} $\tau_\text{c} < c_\text{sim}$}
            \State Add $(\tilde{e},r_\text{sim},m_\text{sim},c_\text{sim})$ to $\mathcal{E}'$
          \EndIf
      \EndFor
      \State Sort $\mathcal{E}'$ by $r_\text{sim},m_\text{sim},c_\text{sim}$
      \For{$\mathbf{e}$ in $\mathcal{E}'$}
        \State Add $\mathbf{e}[0][\text{mentions}]$ to $\mathcal{E}$
      \EndFor
      \State Drop any set in $\mathcal{E}$ that is a subset of another set in $\mathcal{E}$
      \State \Return $\mathcal{E}$
      \EndFunction
  \end{algorithmic}
  \label{algo:getAlternatives}
\end{algorithm}
\section{Extra Counterfactual Examples}
\label{apdx:extra_examples}

In Figure \ref{fig:cf_example_others}, we demonstrated three other examples of a factual bias failure of a DocRE model. Some examples also include relations that are spanning across multiple sentences which a DocRE model should be capable to extract. However, after entity replacement the model (which is trained only on factual data) only manages to predict on the original set.

\end{document}

%% file: figures/CF_example_1.tex
\begin{figure}[t]
    \centering
    \includegraphics[width=\linewidth, trim=0 0 0 0, clip] {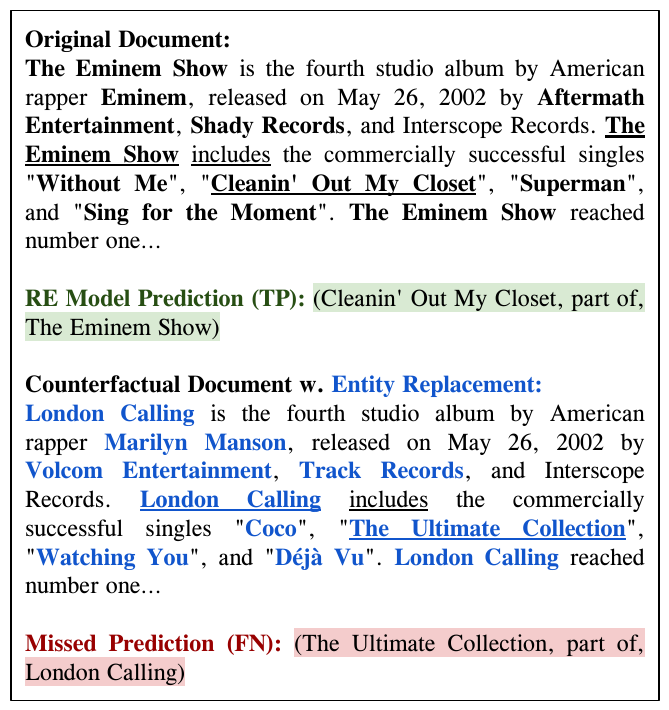}
    \caption{Document from Re-DocRED \citep{tan-etal-2022-revisiting} and counterfactual version  generated with entity replacement. 
    A model trained on factual data extracts the
    original triple,  but fails on its counterfactual (CF) counterpart.
    Thus, the model is relying on spurious
    patterns such as entity biases.
We address this  by generating CF data and training RE
models on them.
    }
    \label{fig:cf_example_main}
\end{figure}


%% file: tables/results.tex
\begin{table}[!t]
    \small
	\setlength\tabcolsep{5pt}
	\centering
    \begin{tabular}{l@{\hspace{20pt}}|ccc|c}
        \toprule
        \textbf{Training Data} & \textbf{{PRC}} & \textbf{REC} & \textbf{F1} & \textbf{CONS}\\
        \midrule
        Re-DocRED & \textbf{88.1} & \textbf{69.8} & \textbf{78.0} & \colorbox{pink}{68.6} \\
        \midrule
        CF Only & 85.3 & 62.8 & 72.4 & \textbf{89.5} \\
        Re-DocRED + CF & \textit{85.7} & \textit{68.8} & \textit{76.3} & \underline{\textit{88.3}} \\
        \bottomrule
    \end{tabular}
    \caption{Evaluation results on
      factual (Re-DocRED), counterfactual (CF Only) and
      combined (Re-DocRED + CF) data.
    Our measures are Precision (PRC),
    Recall (REC) and F1 score on Re-DocRED's test set. Using a counterfactual counterpart of the test set,
    we report consistency (CONS) results of each approach. (All reported numbers are the median over 5 runs with different random seeds.)}
	\label{tab:exp_results}
\end{table}

%% file: figures/CF_example_2.tex
\begin{figure}[!t]
    \centering
    \includegraphics[width=\linewidth, trim=0 0 0 0, clip] {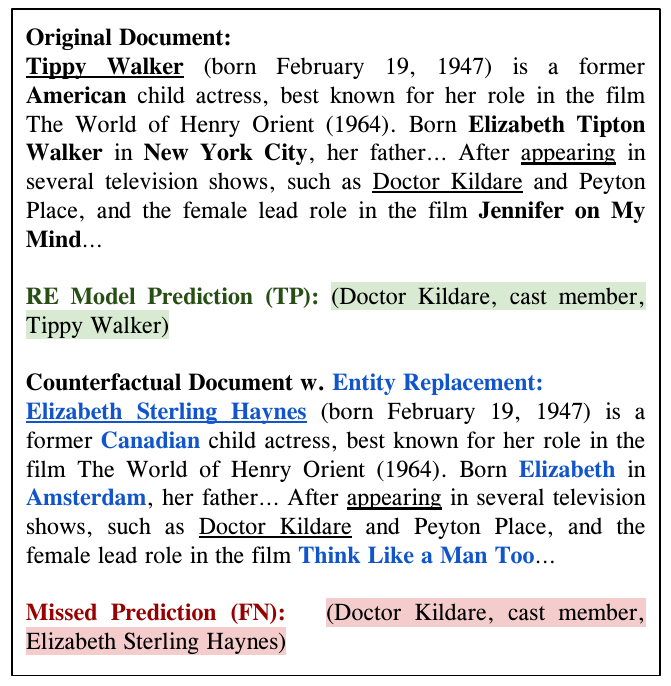}
    \includegraphics[width=\linewidth, trim=0 0 0 0, clip] {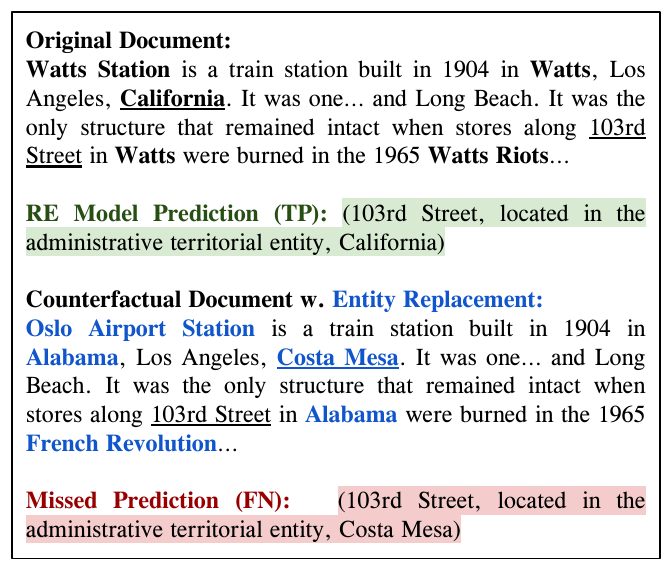}
    \includegraphics[width=\linewidth, trim=0 0 0 0, clip] {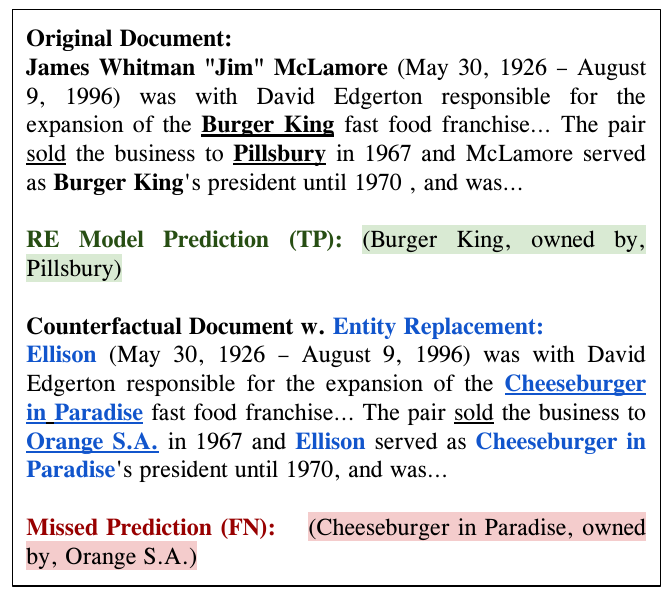}
    \caption{Three other examples of original documents and their counterfactual counterparts.
    In all three we observe a failure in predicting the counterfactual, while all information required
    for the relation to be extracted are present (\underline{Underlined}).}
    \label{fig:cf_example_others}
\end{figure}